%% file: iclr2027_conference.tex
\documentclass{article} 
\usepackage{iclr2027_conference,times}

\input{math_commands.tex}

\usepackage{hyperref}
\usepackage{url}
\usepackage{amsmath,amssymb,amsthm,mathtools}
\usepackage{algorithm}
\usepackage{algorithmic}
\usepackage{float}

\usepackage{booktabs}

\newtheorem{theorem}{Theorem}[section]

\newtheorem{proposition}[theorem]{Proposition}
\newtheorem{assumption}[theorem]{Assumption}

\title{RBGA: Residual-Budget Guide Arbitration for Offline-to-Online Safe Reinforcement Learning in Dexterous Manipulation}

\author{%
\parbox[t]{\dimexpr\textwidth-2\tabcolsep\relax}{%
\centering
\normalfont
\mbox{\textbf{Zongyou Yu}$^{1,2}$}\hfill
\mbox{\textbf{Ao Zhou}$^{1}$}\hfill
\mbox{\textbf{Xiongbo Shi}$^{1}$}\hfill
\mbox{\textbf{Jiayi Guan}$^{3}$}\hfill
\mbox{\textbf{Shihang Du}$^{1}$}\hfill
\mbox{\textbf{Xudong Zhang}$^{1}$}\\[0.2em]
\textbf{Shangding Gu}$^{4,5}$ \quad
\textbf{Junqiao Zhao}$^{1}$ \quad
\textbf{Hanli Wang}$^{1}$ \quad
\textbf{Guang Chen}$^{1,2}$%
\thanks{Corresponding author: Guang Chen
(\texttt{guangchen@tongji.edu.cn}).}\\[0.6em]
$^{1}$Tongji University \\
$^{2}$Shanghai Innovation Institute \\
$^{3}$Xiaomi EV \\
$^{4}$Shanghai Jiao Tong University \\
$^{5}$University of California, Berkeley
}%
}

\iclrfinalcopy 
\begin{document}

\maketitle

\fancyhead{}
\renewcommand{\headrulewidth}{0pt}
 
\begin{abstract}
Dexterous manipulation requires robots to leverage existing datasets for rapid skill acquisition while maintaining safety during real-world deployment to prevent damage to the robot or its surroundings. Offline-to-online safe reinforcement learning (O2O safe RL) provides a promising solution for this problem by leveraging prior experience for online improvement under safety constraints, thereby mitigating the sample inefficiency and exploration risk of learning safe policies from scratch. However, existing O2O safe RL approaches do not explicitly account for the safety budget already consumed when making decisions within an episode. Excessive cost incurred early can therefore leave little room for subsequent exploration and policy improvement. To address this limitation, we propose \textbf{R}esidual-\textbf{B}udget \textbf{G}uide \textbf{A}rbitration (RBGA), an algorithm that explicitly tracks the remaining episode safety budget and uses it to adaptively select between offline guidance and the evolving online policy. At each step, RBGA compares the cost-to-go estimates of the guide and online actions against the remaining budget, enabling step-wise action arbitration throughout the episode. We provide theoretical analysis characterizing the safety implications of residual-budget-based selection and its effect on guide-assisted exploration. Extensive experiments demonstrate that RBGA enables effective online policy improvement while maintaining strong safety performance.  Our code is available at this \href{https://github.com/sii-pilab}{link}.
\end{abstract}


\section{Introduction}
\label{sec:introduction}

Dexterous manipulation enables multi-fingered robotic hands to perform a wide range of contact-rich tasks, including in-hand manipulation, grasping, and tool use~\citep{rajeswaran2018learning,openai2020dexterous,yuan2025crossdex}. Despite substantial progress, improving the safety of dexterous manipulation remains a pressing challenge~\citep{garcia2015survey,jain2023constrained,geng2025redman}. Safe reinforcement learning (RL) addresses this challenge by optimizing task performance subject to explicit cost constraints~\citep{chow2018lyapunov,tessler2019rcpo,stooke2020pid,satija2020backward}. However, learning a safe dexterous policy from scratch is often prohibitive, as it requires extensive environment interactions that are both costly and risky~\citep{ray2019benchmarking,thananjeyan2021recovery,dulacarnold2021challenges}. Offline safe RL reduces this burden by learning policies from a fixed dataset without additional online interaction~\citep{xu2022cpq,lee2022coptidice,liu2023cdt}. Yet its performance is constrained by the coverage of the offline data~\citep{levine2020offline,liu2024datasets}, making learned policies vulnerable to distribution shift and out-of-distribution (OOD) actions when encountering unseen states~\citep{fujimoto2019bcq,kumar2019bear,kumar2020cql}. Offline-to-online (O2O) safe RL therefore offers an appealing alternative by leveraging policies learned offline to accelerate online adaptation, substantially reducing the extensive environment interactions required to learn policies from scratch.

Compared with O2O RL, O2O safe RL remains relatively underexplored \citep{li2024gold,cao2025fosp,chen2026marvel}. Existing policy-finetuning methods continue online learning from an offline-pretrained policy, although this transfer can suffer from mismatched value estimates and Lagrange multipliers between offline and online learning \citep{chen2026marvel}. To facilitate safer and more effective adaptation, some works instead use an offline policy to guide the learning of a separate online policy \citep{li2024gold}, while others preserve offline knowledge through model-based fine-tuning and safe policy expansion \citep{cao2025fosp}. Despite these advances, existing methods do not explicitly account for the cumulative cost already incurred in the current episode when making online decisions. This limitation is particularly important in dexterous manipulation, where contact-rich interactions can produce abrupt safety violations and nonuniform cost accumulation over an episode. Therefore, developing a dedicated O2O safe RL algorithm that explicitly accounts for within-episode cost accumulation is particularly important for dexterous manipulation.

In this work, to address the lack of trajectory-level budget awareness in O2O safe RL, we propose Residual-Budget Guide Arbitration (RBGA), an algorithm that explicitly incorporates the remaining episode cost budget into online policy learning. First, RBGA tracks the remaining cost budget along each trajectory to characterize the current safety margin throughout online interaction. Then, at each interaction step, it evaluates the actions proposed by the offline guide and the online policy based on their estimated cost-to-go, and adaptively selects between them according to the remaining budget. This step-wise arbitration allows the online policy to take control when the guide action becomes unfavorable under the current budget, while retaining useful offline guidance when it remains admissible. Finally, extensive experiments on Redman~\citep{geng2025redman} demonstrate the effectiveness of RBGA across a range of dexterous manipulation tasks. The main contributions of this work are summarized as follows:

\begin{itemize}
    \item To the best of our knowledge, we are the first to explicitly incorporate trajectory-level episode cost budget information into O2O safe RL for dexterous manipulation.
    \item We propose RBGA, a step-wise action arbitration algorithm that adaptively balances offline guidance and the evolving online policy based on the remaining episode budget and estimated cost-to-go. 
    \item Extensive experiments across diverse dexterous manipulation tasks demonstrate the effectiveness and advantages of RBGA, showing strong policy performance while maintaining favorable safety behavior during online adaptation.
\end{itemize}

\section{Related Work}
\label{sec:related_work}

\textbf{RL for Dexterous Manipulation.}
RL has enabled multi-fingered robotic hands to acquire complex manipulation skills from interaction, with early studies demonstrating effective learning from demonstrations and large-scale simulation \citep{rajeswaran2018learning,zhu2019dexterous,openai2020dexterous}. Subsequent work has improved the applicability of learned policies through model-based planning and more efficient policy optimization \citep{nagabandi2020deep,charlesworth2021solving}. Recent studies further extend dexterous RL to unseen objects and real-world manipulation \citep{chen2022system,qin2023dexpoint,qi2023hora,qi2023rotateit}, as well as more challenging bimanual tasks \citep{chen2022bidexhands,huang2023dynamic}. Despite these advances, most prior work considers dexterous manipulation without explicit safety constraints. Safety is nevertheless critical for dexterous manipulation, as unsafe actions during contact-rich interaction may damage the robotic hand or its surroundings. Research on safe dexterous manipulation remains limited. A few studies incorporate explicit constraints into dexterous policy learning or extend safe RL to bimanual manipulation \citep{jain2023constrained,zhan2024safe}. ReDMan further establishes a dedicated benchmark for evaluating safe RL algorithms on dexterous manipulation tasks \citep{geng2025redman}. These works mainly learn or evaluate safe policies through online interaction, leaving how offline experience can facilitate subsequent safe adaptation largely unexplored.

\textbf{Offline-to-Online RL.}
Offline-to-online (O2O) RL leverages offline pretraining to accelerate online policy improvement. A natural strategy is to initialize online learning with an offline-trained policy and continue fine-tuning it using collected experience \citep{nair2020awac}. To improve the stability and efficiency of this transition, some methods balance offline and online samples during replay \citep{lee2022balanced,ball2023rlpd}, while others correct value estimates inherited from offline training before or during online fine-tuning \citep{yu2023alignment,nakamoto2023calql,zhang2024so2,luo2024ocrcft}. Instead of directly fine-tuning the offline policy, JSRL uses a fixed guide policy to facilitate exploration through progressively shortened roll-ins \citep{uchendu2023jumpstart}, whereas PEX retains separate offline and online policies and selects between their proposed actions according to estimated values \citep{zhang2023pex}. Despite substantial progress, most O2O RL methods are developed for unconstrained reward maximization and do not address safety constraints during online adaptation. Recent O2O safe RL studies begin to address this problem by incorporating safety into offline-to-online transfer. GOLD uses an offline sequence policy to guide the learning of an online safe policy \citep{li2024gold}. However, GOLD relies on a Decision Transformer as the offline guide and performs guided online distillation, differing from the matched offline-policy setting considered in this work. FOSP combines offline model learning with safe policy expansion for online fine-tuning \citep{cao2025fosp}. However, FOSP adopts a model-based safe O2O setting with a learned world model, which is not directly comparable to the model-free setting considered in this work.

Despite these advances, existing O2O safe RL studies have not addressed dexterous manipulation, where unsafe exploratory actions can damage the robotic hand or its surroundings. Learning a safe dexterous policy from scratch requires costly and risky online interaction, while offline safe RL remains limited by the coverage of the fixed dataset and degrades under distribution shift. O2O safe RL is therefore particularly appealing for dexterous manipulation, as it reuses offline knowledge to accelerate online adaptation under safety constraints. To the best of our knowledge, we are the first to study O2O safe RL for dexterous manipulation.

\section{Preliminaries}
\label{sec:preliminaries}

This section introduces the constrained RL formulation and the O2O safe RL setting used throughout the paper. We first describe finite-horizon Markov decision processes and their extension to explicit cost constraints. We then specify the offline and online policies considered by RBGA.

\subsection{Constrained Markov Decision Processes}
\label{sec:cmdp}

Standard RL is formulated as a finite-horizon Markov decision process (MDP) $\mathcal M_{\mathrm{MDP}}=(\mathcal S,\mathcal A,P,r,\rho_0,\gamma,H)$, where $\mathcal S$ and $\mathcal A$ are the state and action spaces, $P(\cdot\mid s,a)$ is the transition kernel, $r:\mathcal S\times\mathcal A\rightarrow\mathbb R$ is a bounded reward function, $\rho_0$ is the initial-state distribution, $\gamma\in(0,1]$ is the discount factor, and $H$ is the finite episode horizon. A policy $\pi$ induces a trajectory through $s_0\sim\rho_0$, $a_t\sim\pi(\cdot\mid s_t)$, and $s_{t+1}\sim P(\cdot\mid s_t,a_t)$. For policies that depend on the episode step, we omit the time index for simplicity. Standard RL maximizes the expected discounted reward.

Safe RL augments the MDP with a bounded non-negative cost function $c:\mathcal S\times\mathcal A\rightarrow\mathbb R$ and a cost limit $d$, giving the constrained MDP $\mathcal M=(\mathcal S,\mathcal A,P,r,c,\rho_0,\gamma,H,d)$ \citep{altman1999cmdp,achiam2017cpo}. For any policy $\pi$, the expected reward and cost are
\begin{equation}
J_R(\pi)=\mathbb E_{\tau\sim\pi}\left[\sum_{t=0}^{H-1}\gamma^t r(s_t,a_t)\right],
\qquad
J_C(\pi)=\mathbb E_{\tau\sim\pi}\left[\sum_{t=0}^{H-1}\gamma^t c(s_t,a_t)\right],
\label{eq:policy_returns}
\end{equation}
where $\tau$ denotes a trajectory induced by policy $\pi$, and $R$ and $C$ denote reward and cost, respectively.

The common objective of constrained RL is to maximize the cumulative reward while satisfying the cumulative cost constraint. Accordingly, the optimal feasible policy $\pi^\star$ is defined as
\begin{equation}
\pi^\star\in\operatorname*{arg\,max}_{\pi}J_R(\pi)
\quad\text{s.t.}\quad J_C(\pi)\leq d.
\label{eq:cmdp_objective}
\end{equation}
The objective maximizes task return while constraining the expected cumulative cost. In dexterous manipulation, the state represents the hand--object configuration, the action specifies a control command for the robotic hand, and the reward measures progress toward the manipulation objective, while the cost function penalizes unsafe behavior during interaction.

\subsection{Offline-to-Online Safe Reinforcement Learning}
\label{sec:safe_o2o}

Offline-to-online (O2O) RL leverages offline data or a previously learned policy to support subsequent online policy learning \citep{uchendu2023jumpstart,zhang2023pex}, while O2O safe RL further requires this online improvement to satisfy the cost constraint in Eq.~(\ref{eq:cmdp_objective}) \citep{li2024gold,cao2025fosp,chen2026marvel}. In O2O safe RL, an offline policy $\pi_g$ is first trained on a fixed offline dataset $\mathcal D_{\mathrm{off}}$ of transitions $(s,a,r,c,s',\delta)$, where $r=r(s,a)$ and $c=c(s,a)$ are the observed reward and cost, $s'$ is the next state, and $\delta\in\{0,1\}$ indicates episode termination. The learned offline policy is then used to support subsequent online policy learning. We denote the online policy by $\pi_o$ and update it from newly collected transitions using a constrained RL algorithm.

\section{Methodology}
\label{sec:method}

In this section, we first introduce Residual-Budget Guide Arbitration (RBGA) for O2O safe RL in dexterous manipulation. We first describe how guide horizon scheduling and residual-budget-based action selection combine offline guidance with online learning. We then provide theoretical analysis of the cost of the composite policy and the reward performance of the learned online policy.

\subsection{Residual-Budget Guide Arbitration}
\label{sec:rbga}

\textbf{Guide horizon scheduling.} Inspired by JSRL~\citep{uchendu2023jumpstart}, RBGA combines a fixed offline guide policy $\pi_g$ and a trainable online policy $\pi_o$ into a composite policy $\pi$ for online interaction. Consider a dexterous manipulation episode of at most $H$ control steps and a total of $N$ online training iterations. To limit dependence on the offline guide, we specify a guidance cutoff $N_g\leq N$. During the first $N_g$ training iterations, the offline guide participates in action selection, while the online policy acts alone thereafter. The online policy is initialized independently and updated from collected experience. At iteration $i$, the maximum guide horizon $H_i\in\{0,\ldots,H\}$ controls the number of initial episode steps during which guide actions will be considered. Decreasing $H_i$ allows the online policy to take control earlier as training progresses. Specifically, in JSRL, the guide horizon is directly reduced across curriculum stages, introducing abrupt changes in the state distribution encountered by the online policy. To provide a smoother transition from offline guidance to online control, we instead gradually decrease the maximum guide horizon as training progresses and consider two scheduling strategies, \textit{Linear} and \textit{Step}. For the linear schedule, we set
\begin{equation}
H_i=\operatorname{round}\!\left(H_0\left[1-\frac{i}{N_g}\right]_+\right),
\qquad N_g>0,
\label{eq:linear_guide_schedule}
\end{equation}
where $i\in\{0,\ldots,N\}$ counts completed online training iterations, $H_0\leq H$ is the initial maximum guide horizon, $[\cdot]_+$ clips negative values to zero, and $\operatorname{round}(\cdot)$ rounds to the nearest integer. Thus, $H_i=0$ for $i\geq N_g$. A step schedule is described in Appendix~\ref{app:guide_schedule}.

To allow the online policy to take over at different stages of an episode, we sample $h$ uniformly from $\{0,\ldots,H_i\}$ at the beginning of each episode, independently of the initial state. The sampled horizon $h$ remains fixed within that episode, and guide actions are considered only when $t<\min\{h,H_i\}$. Here, $t$ counts control steps within the current manipulation attempt, while $i$ counts training iterations. If an episode spans several updates, the current $H_i$ continues to apply, so a decrease in $H_i$ can end guidance before the sampled horizon $h$ is reached.

\textbf{Residual budget.} A limited guide horizon does not by itself prevent unsafe actions. The offline guide may produce actions that substantially violate safety constraints and potentially damage the robotic hand or its surroundings due to distribution shift between offline training and online interaction. We therefore account for the costs already incurred when selecting actions within the guide horizon. Before executing the action at step $t$, we define the residual budget as
\begin{equation}
b_t=d-\sum_{j=0}^{t-1}\gamma^j c_j,\qquad b_0=d,
\label{eq:remaining_budget}
\end{equation}
where $d$ is the cost limit for a complete manipulation episode, $c_j=c(s_j,a_j)$ measures the safety cost incurred by the hand at step $j$, and $\gamma$ is the discount factor. In dexterous tasks, the cost penalizes unsafe behaviors of the robotic hand during manipulation. Thus, $b_t$ is the budget left for the rest of the manipulation, and $b_t<0$ means that the accumulated discounted cost has already exceeded $d$.

To select an action under the residual budget, we first obtain candidate actions from the two policies. Given the current state $s_t$, we sample an online candidate $a_t^o\sim\pi_o(\cdot\mid s_t)$ and, when $t<\min\{h,H_i\}$, a guide candidate $a_t^g\sim\pi_g(\cdot\mid s_t)$. To compare the candidates against the residual budget, we evaluate the expected discounted cost of executing each candidate and following the current online policy thereafter. For a fixed $\pi_o$, this cost-to-go is defined as:
\begin{equation}
Q_C^{\pi_o}(s_t,a_t)=\mathbb E_{\pi_o}\left[\sum_{j=t}^{H-1}\gamma^{j-t}c_j\,\middle|\,s_t,a_t\right],
\label{eq:cost_to_go}
\end{equation}
where $a_t$ is the candidate control command being evaluated and subsequent actions follow $\pi_o$. For notational simplicity, we omit the time index on finite-horizon value and advantage functions when the corresponding episode step is clear from context. Since this expected cost is not directly available, we learn an auxiliary cost estimator $\widehat Q_C^{\pi_o}(s,a)$ from online experience to approximate Eq.~(\ref{eq:cost_to_go}) and use it to evaluate both candidates, as detailed in Appendix~\ref{app:aux_cost_estimator}. For comparison with $b_t$, we define $\widehat q_t^g=\gamma^t\widehat Q_C^{\pi_o}(s_t,a_t^g)$ and $\widehat q_t^o=\gamma^t\widehat Q_C^{\pi_o}(s_t,a_t^o)$. These estimates express the candidates' remaining costs on the same discount scale as the episode budget.

\textbf{Step-wise arbitration.}  RBGA combines the guide horizon, residual budget, and estimated cost-to-go to select the executed action. To retain useful guidance while limiting its additional cost, we introduce a comparison margin $\zeta\geq0$, which bounds the additional estimated cost-to-go of the guide candidate relative to the online candidate. The executed action is then selected as
\begin{equation}
a_t=\begin{cases}
a_t^o, & t\geq\min\{h,H_i\},\\
a_t^g, & t<\min\{h,H_i\}\ \text{and}\ \widehat q_t^g\leq\min\{b_t,\widehat q_t^o+\gamma^t\zeta\},\\
a_t^o, & t<\min\{h,H_i\},\ \widehat q_t^o\leq b_t\ \text{and}\ \widehat q_t^g>\min\{b_t,\widehat q_t^o+\gamma^t\zeta\},\\
\operatorname*{arg\,min}_{a\in\{a_t^g,a_t^o\}}\widehat Q_C^{\pi_o}(s_t,a), & \text{otherwise}.
\end{cases}
\label{eq:rbga_selection}
\end{equation}
When $t\geq\min\{h,H_i\}$, the guide is no longer eligible, and RBGA executes the online candidate $a_t^o$. When $t<\min\{h,H_i\}$ and $\widehat q_t^g\leq\min\{b_t,\widehat q_t^o+\gamma^t\zeta\}$, the guide remains eligible, its estimated cost-to-go satisfies the remaining budget, and its additional estimated cost relative to the online candidate is within the comparison margin; RBGA therefore selects $a_t^g$. When $t<\min\{h,H_i\}$, $\widehat q_t^o\leq b_t$, and $\widehat q_t^g>\min\{b_t,\widehat q_t^o+\gamma^t\zeta\}$, the online candidate satisfies the remaining budget while the guide fails either the budget or comparison-margin condition, so RBGA selects $a_t^o$. Otherwise, both candidates exceed the remaining budget, and RBGA selects the candidate with the smaller estimated cost-to-go. We provide a detailed description of RBGA in Algorithm~\ref{alg:rbga}.

\begin{algorithm}[t]
\caption{Residual-Budget Guide Arbitration (RBGA)}
\label{alg:rbga}

\begin{algorithmic}[1]
\STATE \textbf{Input:} fixed guide policy $\pi_g$, task horizon $H$, total training iterations $N$, guide horizon schedule with cutoff $N_g$, cost limit $d$, discount factor $\gamma$, and comparison margin $\zeta$.
\STATE Initialize the online learner with policy $\pi_o$ and the auxiliary estimator $\widehat Q_C^{\pi_o}$ independently of $\pi_g$; set $\mathcal B\leftarrow\varnothing$ and $i\leftarrow0$.
\STATE Reset the environment; set $t\leftarrow0$, $b_t\leftarrow d$, and sample $h\sim\operatorname{Unif}\{0,\ldots,H_i\}$.
\WHILE{$i<N$}
    \FOR{each environment step}
        \STATE Select $a_t$ using Eq.~(\ref{eq:rbga_selection}) after warm-up; use the warm-up procedure otherwise.
        \STATE Execute $a_t$; observe $r_t,c_t,s_{t+1},\delta_t$; append the transition to $\mathcal B$.
        \STATE Set $b_{t+1}\leftarrow b_t-\gamma^t c_t$ and $t\leftarrow t+1$.
        \IF{the episode has ended}
            \STATE Reset the environment; set $t\leftarrow0$, $b_t\leftarrow d$, and sample a new $h\sim\operatorname{Unif}\{0,\ldots,H_i\}$.
        \ENDIF
    \ENDFOR
    \STATE $\pi_o,\widehat Q_C^{\pi_o},i\leftarrow\textsc{TrainPolicy}(\pi_o,\widehat Q_C^{\pi_o},\mathcal B,i,N)$; obtain $H_i$ from the schedule.
\ENDWHILE
\STATE \textbf{Return:} the learned online policy $\pi_o$.
\end{algorithmic}

\end{algorithm}

\subsection{Theoretical Analysis}
\label{sec:theoretical_analysis}

In this section, we analyze the cost of RBGA action selection and the reward performance of the learned policy. We first derive local cost bounds from the remaining budget and cost estimates, and then obtain an expected cost bound for the composite policy. Subsequently, we analyze the reward gap relative to the optimal feasible policy under state coverage and policy advantage conditions. To account for the accuracy of the cost estimates in Eq.~(\ref{eq:rbga_selection}), we introduce the following assumption.

\begin{assumption}[Cost estimation error]
\label{ass:cost_error}
For a fixed online policy $\pi_o$ and cost estimator $\widehat Q_C^{\pi_o}$, there exists a finite constant $\varepsilon_C\geq0$ such that, at each time step $t<\min\{h,H_i\}$, the cost estimates of both candidate actions in Eq.~(\ref{eq:rbga_selection}) satisfy:
\begin{equation}
\left|\widehat Q_C^{\pi_o}(s_t,a)-Q_{C}^{\pi_o}(s_t,a)\right|\leq\varepsilon_C,\qquad\forall a\in\{a_t^g,a_t^o\},
\label{eq:cost_error_bound}
\end{equation}
where $\varepsilon_C$ is a uniform bound on their estimation errors.
\end{assumption}

Based on Assumption~\ref{ass:cost_error} and the remaining budget in Eq.~(\ref{eq:remaining_budget}), we derive the cost bounds for a selected action, as shown in Proposition~\ref{prop:local_cost_bounds}. Proofs and discussions are in Appendix~\ref{app:proof_local_cost_bounds}.

\begin{proposition}[Local cost bounds]
\label{prop:local_cost_bounds}
Under Assumption~\ref{ass:cost_error}, consider a time step $t<\min\{h,H_i\}$. For a selected action $a_t$ satisfying $\gamma^t\widehat Q_C^{\pi_o}(s_t,a_t)\leq b_t$, the accumulated discounted cost plus the expected discounted cost of executing $a_t$ and following $\pi_o$ thereafter satisfies:
\begin{equation}
\sum_{j=0}^{t-1}\gamma^j c_j+\gamma^t Q_{C}^{\pi_o}(s_t,a_t)
\leq d+\gamma^t\varepsilon_C\leq d+\varepsilon_C.
\label{eq:local_budget_bound}
\end{equation}
For every guide action selected by Eq.~(\ref{eq:rbga_selection}), the difference between its cost-to-go and that of the online candidate satisfies:
\begin{equation}
Q_{C}^{\pi_o}(s_t,a_t^g)-Q_{C}^{\pi_o}(s_t,a_t^o)
\leq\zeta+2\varepsilon_C.
\label{eq:guide_selection_cost_bound}
\end{equation}
\end{proposition}

Proposition~\ref{prop:local_cost_bounds} bounds the accumulated discounted cost together with the expected continuation cost for actions satisfying the budget condition, and the additional cost-to-go of a guide action relative to the online candidate. By combining Eq.~(\ref{eq:guide_selection_cost_bound}) with the finite-horizon cost form of the performance difference lemma \citep{kakade2002approximately}, we obtain the episode cost bound in Proposition~\ref{thm:global_cost_bound}. 


\begin{proposition}[Cost bound for the composite policy]
\label{thm:global_cost_bound}
Under Assumption~\ref{ass:cost_error}, fix $\pi_o$, $\widehat Q_C^{\pi_o}$, and $H_i$ throughout an episode, and sample $h\sim\operatorname{Unif}\{0,\ldots,H_i\}$ independently of the initial state. The composite policy $\pi$ defined by Eq.~(\ref{eq:rbga_selection}) satisfies:
\begin{equation}
[J_C(\pi)-d]_+ \leq[J_C(\pi_o)-d]_+ +(\zeta+2\varepsilon_C)\sum_{t=0}^{H_i-1}\gamma^t\frac{H_i-t}{H_i+1},
\label{eq:global_constraint_violation_bound}
\end{equation}
where $[u]_+=\max\{u,0\}$, $J_C(\pi)$ is the expected discounted episode cost under the composite policy, with expectation over the initial state, sampled guide horizon, candidate actions, and state transitions, and $(H_i-t)/(H_i+1)=\Pr(h>t)$ is the probability that the sampled guide horizon extends beyond step $t$ under uniform sampling, for $0\leq t<H_i$.

\end{proposition}

Proposition~\ref{thm:global_cost_bound} bounds the additional violation of the expected cost constraint in terms of the comparison margin, estimation error, and maximum guide horizon. The final term does not increase as the maximum guide horizon is reduced and equals zero when the guide is removed. Proposition~\ref{prop:changing_policy_cost_bound} in Appendix~\ref{app:proof_global_cost_bound} extends the analysis to policy updates within an episode.

To analyze reward performance under RBGA roll-ins, we consider the state distributions used for policy learning, following the state coverage analysis of guide-based policy search \citep{uchendu2023jumpstart}. For a learned online policy $\pi_o$, the reward advantage at time $t$ is defined as:
\begin{equation}
A_{R}^{\pi_o}(s,a)
=Q_{R}^{\pi_o}(s,a)
-\mathbb E_{a'\sim\pi_o(\cdot\mid s)}[Q_{R}^{\pi_o}(s,a')],
\label{eq:reward_advantage}
\end{equation}
where $Q_{R}^{\pi_o}(s,a)$ is the expected discounted reward $\sum_{j=t}^{H-1}\gamma^{j-t}r_j$ obtained by executing $a$ in $s$ and following $\pi_o$ thereafter, and $r_j=r(s_j,a_j)$. Based on this advantage function, we introduce conditions on state coverage and the remaining advantage of the optimal feasible policy.

\begin{assumption}[State coverage and policy advantage]
\label{ass:rollin_learning}
For the learned online policy $\pi_o$, there exist a finite coverage coefficient $D_\infty\geq1$ and finite reward advantage bounds $\varepsilon_{R,t}\geq0$ such that, at every time step $t$, the state distributions satisfy:
\begin{equation}
\rho_t^{\pi^\star}\leq D_\infty\mu_t,
\label{eq:rbga_rollin_coverage}
\end{equation}
where $\mu_t$ is the marginal state distribution at time $t$ induced by RBGA roll-ins, $\rho_t^{\pi^\star}$ is the corresponding distribution under $\pi^\star$, and $D_\infty$ bounds their relative probabilities. Under the roll-in distribution, the expected positive advantage of the optimal feasible policy satisfies:
\begin{equation}
\mathbb E_{s\sim\mu_t}
\left[\left[\mathbb E_{a\sim\pi^\star(\cdot\mid s)}
A_{R}^{\pi_o}(s,a)\right]_+\right]
\leq\varepsilon_{R,t},
\label{eq:rollin_advantage_error}
\end{equation}
where $\varepsilon_{R,t}$ bounds the remaining positive reward advantage of $\pi^\star$ under the roll-in distribution.
\end{assumption}

By applying the state coverage condition to the positive advantage in Eq.~(\ref{eq:rollin_advantage_error}), we obtain the reward performance bound shown in Proposition~\ref{thm:screened_exploration}. Proofs and discussions are in Appendix~\ref{app:policy_search_rollins} and~\ref{app:proof_reward_bound}.

\begin{proposition}[Reward performance]
\label{thm:screened_exploration}
Under Assumption~\ref{ass:rollin_learning}, the reward gap between the learned online policy $\pi_o$ and the optimal feasible policy $\pi^\star$ satisfies:
\begin{equation}
J_R(\pi^\star)-J_R(\pi_o)
\leq D_\infty\sum_{t=0}^{H-1}\gamma^t\varepsilon_{R,t}.
\label{eq:reward_performance_bound}
\end{equation}
\end{proposition}

Proposition~\ref{thm:screened_exploration} bounds the reward gap between the learned online policy $\pi_o$ and the optimal feasible policy $\pi^\star$. Once $H_i=0$, Eq.~(\ref{eq:rbga_selection}) gives $\pi=\pi_o$. The same bound therefore applies to the composite policy after the guide is removed, provided that the learned online policy satisfies Assumption~\ref{ass:rollin_learning}.

\begin{figure}[t]
    \centering
    \includegraphics[width=\linewidth]{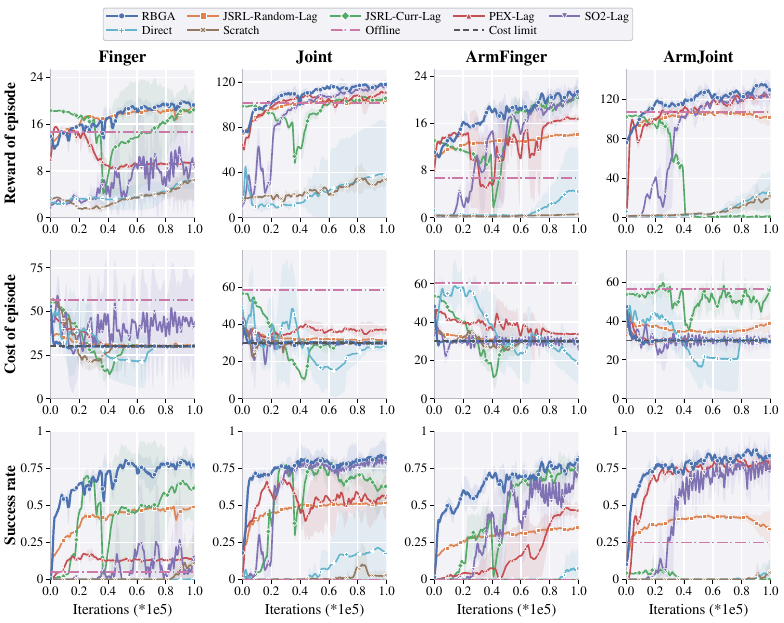}
    \caption{The figure depicts the reward, cost, and success rate curves of RBGA and the baseline algorithms in the comparison experiment. The shaded areas represent variance over three random seeds. The dashed line indicates the episode cost limit $d=30$.}
    \label{fig:main_results}
\end{figure}

\section{Experiment}
In this section, we conduct comprehensive comparative experiments between RBGA and prior O2O safe RL methods in dexterous manipulation tasks. Moreover, we conduct ablation studies to evaluate the effectiveness of guide horizon scheduling and the contribution of residual-budget arbitration.

\subsection{Task and Baseline}
\textbf{Task.} We evaluate RBGA on four dexterous manipulation tasks from the ReDMan benchmark~\citep{geng2025redman}, including \emph{Joint}, \emph{Finger}, \emph{ArmJoint}, and \emph{ArmFinger}. These tasks combine two bimanual object-transfer settings with constraints on either a single joint or multiple finger joints; task descriptions are provided in Appendix~\ref{app:task_descriptions}. The original ReDMan environments use binary cost signals to indicate safety violations, resulting in abrupt changes between safe and unsafe states. To provide a more informative safety signal for policy learning, we replace the binary cost with a continuous cost while preserving the original safety definition of each task. The resulting continuous cost signal $c_t=c(s_t,a_t)$ is used to compute the cumulative cost over an episode. We set the episode cost limit to $d=30$ for all tasks.

\textbf{Baselines.}
We compare RBGA with the following six baselines:
\textit{(i)} \textbf{JSRL-Random-Lag} \citep{uchendu2023jumpstart}: the random-switching variant of JSRL, which uniformly samples the number of initial guide steps for each episode and then switches from the offline guide to the online policy;
\textit{(ii)} \textbf{JSRL-Curr-Lag} \citep{uchendu2023jumpstart}: the curriculum variant of JSRL, which progressively reduces the number of initial guide steps during training, allowing the online policy to take control earlier in each episode;
\textit{(iii)} \textbf{PEX-Lag} \citep{zhang2023pex}: adaptively selects between actions proposed by a fixed offline policy and a trainable online policy using reward value estimates with a Lagrangian cost penalty;
\textit{(iv)} \textbf{SO2-Lag} \citep{zhang2024so2}: fine-tunes an offline-pretrained policy using a mixture of offline and online replay data, perturbed target actions, and more frequent critic updates;
\textit{(v)} \textbf{Direct}: initializes the online actor and the reward and cost critics from an offline-pretrained checkpoint, and then directly fine-tunes them using SAC-Lag;
\textit{(vi)} \textbf{Scratch}: trains SAC-Lag from scratch using online interactions, without offline pretraining. Since JSRL, PEX, and SO2 were originally developed for unconstrained O2O RL, we adapt them to O2O safe RL by using cost critics and adaptive Lagrange multipliers to incorporate episode cost constraints into online policy learning, following the Lagrangian formulation used in RBGA.
We additionally report \textbf{Offline}, the performance of the offline-pretrained policy without online fine-tuning, as a reference.

\textbf{Implementation Details.} We use IQL as the offline policy and SAC-Lag as the online policy. See the Appendix~\ref{app:experimental_details} for implementation details. Unless otherwise stated, RBGA uses the Linear schedule with $N_g=40{,}000$ iterations. In addition, we evaluate the compatibility of RBGA with DDPG-Lag~\citep{lillicrap2016ddpg,ji2023omnisafe} as an alternative online learner, with detailed results provided in the Appendix~\ref{app:ddpg_results}.

\subsection{Performance Comparison Experiment}

To evaluate the performance of RBGA across various dexterous manipulation tasks, we conduct comprehensive comparative experiments. The results involve comparing RBGA with the baseline algorithms on four tasks: \emph{Finger}, \emph{Joint}, \emph{ArmFinger}, and \emph{ArmJoint}. Figure~\ref{fig:main_results} illustrates the reward, cost, and success rate curves of RBGA and the baseline algorithms on the four tasks. From the illustrated results, it can be observed that, compared with the baseline algorithms, RBGA maintains the episode cost close to the predefined cost limit while achieving higher rewards and success rates. Furthermore, RBGA demonstrates more stable performance compared with the other baseline algorithms during online training. Notably, RBGA achieves the best performance across the four tasks while maintaining favorable safety performance under the cost constraint. The above results indicate that RBGA consistently enables effective online policy improvement across various safe dexterous manipulation tasks while maintaining the episode cost around the predefined cost limit. We also evaluate RBGA on other manipulation tasks in Appendix~\ref{app:additional_tasks}, and describe its deployment for real-world cube grasping in Appendix~\ref{app:real_world}.

\begin{figure}[htbp]
    \centering
    \includegraphics[width=\linewidth]{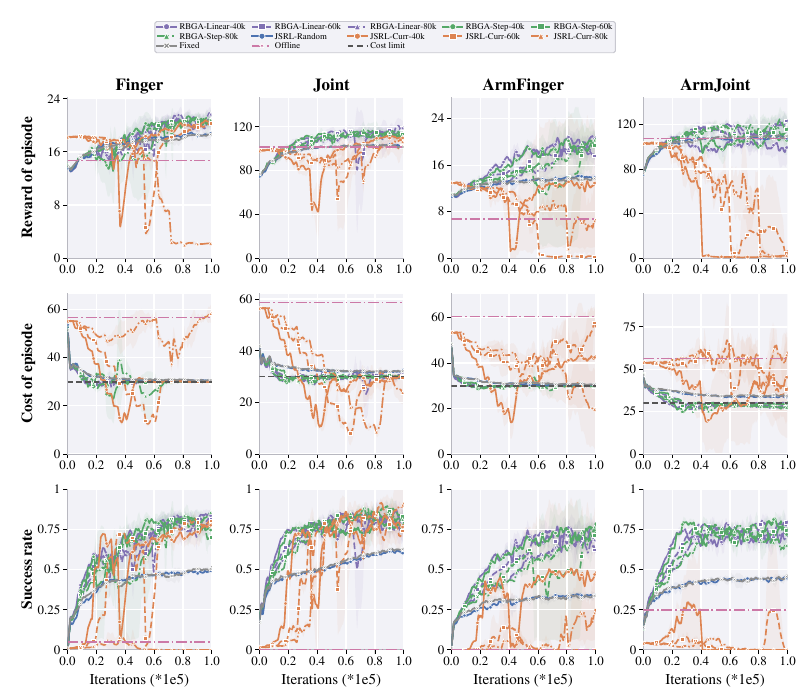}
    \caption{The figure depicts the reward, cost, and success rate curves under different guide horizon scheduling strategies. The shaded areas represent variance over three random seeds. The labels 40k, 60k, and 80k indicate the iterations at which guidance ends, while Random and Fixed retain guidance throughout training. The dashed line indicates the episode cost limit $d=30$.}
    \label{fig:schedule_ablation}
\end{figure}

\subsection{Ablation Study}

To evaluate the effect of different guide horizon scheduling strategies on online learning, we conduct ablation experiments across the four dexterous manipulation tasks. Specifically, we compare the proposed Linear and Step schedules with JSRL-Random and JSRL-Curriculum~\citep{uchendu2023jumpstart}, as well as a Fixed schedule. For Linear, Step, and JSRL-Curriculum, guidance ends at 40k, 60k, or 80k online training iterations, whereas JSRL-Random and Fixed retain guide participation throughout training. Figure~\ref{fig:schedule_ablation} illustrates the reward, cost, and success rate curves of the different scheduling strategies across the four tasks. From the illustrated results, it can be observed that the Linear and Step schedules generally provide stable policy improvement and achieve competitive reward and success rate. Notably, Linear-40k achieves consistently strong performance across the four tasks while maintaining favorable cost control. In contrast, several JSRL-Curriculum configurations exhibit large performance fluctuations or sharp degradation during online training, while JSRL-Random and Fixed also do not consistently match the performance of the decreasing-horizon schedules. These results indicate that sampling the guide horizon within a gradually decreasing maximum horizon provides more stable online learning than using a deterministic stage-wise guide horizon. This is because sampling the guide horizon within the current maximum horizon as it gradually decreases over training allows the online policy to take over at different episode steps, leading to smoother changes in the state distribution and reducing abrupt shifts between curriculum stages. We additionally evaluate the contribution of residual-budget arbitration in Appendix~\ref{app:residual_budget_ablation}.




\section{Conclusion}
In this work, we study O2O safe RL for dexterous manipulation and propose Residual-Budget Guide Arbitration (RBGA). Specifically, we first use a decreasing guide horizon to control the participation of a fixed offline guide during online learning. Based on this, we introduce a residual budget to track the episode cost already incurred and select between the guide and online actions according to the remaining budget and their estimated cost-to-go. Furthermore, we provide theoretical analyses of the cost of the composite policy and the reward performance of the learned online policy. Finally, extensive comparison and ablation experiments on dexterous manipulation tasks demonstrate that RBGA achieves strong task performance while maintaining favorable safety performance.

\section*{AI Assistance Disclosure}

In this work, we used generative AI tools to aid and polish writing, retrieve and discover relevant literature, support research ideation and execution, draft and revise portions of the manuscript, and assist in formulating and checking mathematical claims and proofs. The authors reviewed and verified all AI-assisted content, including literature references, technical descriptions, mathematical derivations, and proofs. We take responsibility for the final content of this work, including text, claims, results, and mathematical statements produced with the aid of generative AI.

\bibliography{iclr2027_conference}
\bibliographystyle{iclr2027_conference}

\newpage

\appendix

\section{Additional Experimental Details}
\label{app:experimental_details}

\subsection{Task Descriptions}
\label{app:task_descriptions}

To evaluate the performance of RBGA in various safe dexterous manipulation tasks with different manipulation settings and joint-range constraints, we select four tasks including \emph{Finger}, \emph{Joint}, \emph{ArmFinger}, and \emph{ArmJoint} from ReDMan~\citep{geng2025redman} as the experimental tasks in this work.

\textbf{Finger.} Two fixed-base dexterous hands cooperate to transfer a ball to a target position. The safety constraint restricts the ranges of motion of three index-finger joints.

\textbf{Joint.} The manipulation objective and hand configuration are the same as in Finger, but the safety constraint applies to a single index-finger joint.

\textbf{ArmFinger.} The object is thrown from a vertically oriented hand and caught by a second hand with its palm facing upward. Multiple index-finger joints are subject to range-of-motion constraints.

\textbf{ArmJoint.} This task uses the same throwing-and-catching setting as ArmFinger, with the safety constraint imposed on a single index-finger joint.

\subsection{Offline Guide and Online Learning}
\label{app:offline_guide}

Unless otherwise stated, RBGA uses implicit Q-learning (IQL) \citep{kostrikov2022iql} to train the offline guide and SAC-Lagrangian (SAC-Lag) \citep{haarnoja2018sac,ji2023omnisafe} for online learning. We train the offline policy $\pi_g$ on a fixed dataset $\mathcal D_{\mathrm{off}}$ containing 1,000,000 transitions per task and keep it fixed during online learning. The critics and the auxiliary cost estimator associated with the online policy $\pi_o$ are initialized independently of the guide policy.

The composite policy $\pi$ collects transitions into the online replay buffer $\mathcal B$. The guide policy remains fixed throughout online learning, whereas the online policy is continually updated using transitions collected in the online replay buffer. Each collection comprises eight steps in each of 512 parallel environments. Learning begins after the buffer contains 500,000 transitions; each subsequent collection is followed by ten training iterations, up to the total of $N$. Each iteration updates the SAC-Lag learner and then the auxiliary estimator using a fresh minibatch. Table~\ref{tab:rbga_hyperparameters} summarizes the architecture and training settings.

\begin{table}[p]
\centering
\caption{RBGA hyperparameters used in the main experiments. The online SAC-Lag cost critic and the auxiliary cost estimator are separate networks.}
\label{tab:rbga_hyperparameters}
\small
\setlength{\tabcolsep}{6pt}
\renewcommand{\arraystretch}{1.08}
\begin{tabular}{@{}p{0.59\linewidth}p{0.35\linewidth}@{}}
\toprule
Parameter & Value \\
\midrule
\multicolumn{2}{@{}l}{\textbf{Network architecture}}\\
Hidden layers of each actor and critic & $[1024,1024,512]$ \\
Hidden activation & ELU \\
Online policy & Tanh-squashed Gaussian \\
Reward critics / SAC-Lag cost critics & 2 / 2 \\
Auxiliary cost critics & 2 \\
\midrule
\multicolumn{2}{@{}l}{\textbf{Online optimization}}\\
Optimizer & Adam \\
Actor / reward critic / cost critic learning rate & $1e-3$ \\
Entropy temperature learning rate & $1e-3$ \\
Initial entropy temperature & $0.01$ \\
Target entropy & Negative action dimension \\
SAC-Lag TD discount & $0.99$ \\
Target soft update coefficient & $0.005$ \\
Maximum gradient norm & $5$ \\
Initial Lagrange multiplier & $0.001$ \\
Lagrange multiplier learning rate & $1e-3$ \\
Lagrange multiplier bounds & $[0,5]$ \\
\midrule
\multicolumn{2}{@{}l}{\textbf{Data collection and replay}}\\
Parallel environments & $512$ \\
Steps per environment per collection & $8$ \\
Training iterations per collection & $10$ \\
Total training iterations $N$ & $100{,}000$ \\
Replay buffer capacity & $1{,}000{,}000$ transitions \\
Minimum replay size before learning & $500{,}000$ transitions \\
Batch size & $4096$ \\
Replay sampling & Uniform \\
\midrule
\multicolumn{2}{@{}l}{\textbf{Guide scheduling and arbitration}}\\
Episode horizon $H$ / initial guide horizon $H_0$ & $75$ / $75$ \\
Guide horizon schedule & Linear \\
Guidance cutoff $N_g$ & $40{,}000$ iterations \\
Episode cost limit $d$ & $30$ \\
Comparison margin $\zeta$ & $1$ \\
Residual-budget / auxiliary TD discount $\gamma$ & $1.0$ \\
Auxiliary estimator warm-up & $2000$ updates \\
\midrule
\multicolumn{2}{@{}l}{\textbf{Auxiliary estimator training}}\\
Learning rate & $1e-3$ \\
Minibatch size & $4096$ \\
Updates per online training iteration & $1$ \\
Sample retention probability & $0.8$ \\
Regression loss & Smooth L1 (threshold $1$) \\
Arbitration estimate & Non-negative target mean \\
Target soft update coefficient & $0.005$ \\
\bottomrule
\end{tabular}
\end{table}

\subsection{Online SAC-Lag Updates}
\label{app:online_training}

\textbf{Critic updates.} For a transition sampled from $\mathcal B$, the reward and cost TD targets are
\begin{equation}
\begin{aligned}
y_R
&=r+\gamma(1-\delta)
\min_{j\in\{1,2\}}Q_{\bar\phi_{R,j}}(s',a')
,\\
y_C
&=c+\gamma(1-\delta)
\max_{j\in\{1,2\}}Q_{\bar\phi_{C,j}}(s',a').
\end{aligned}
\label{eq:sac_td_targets}
\end{equation}
where $(s,a,r,c,s',\delta)$ is a sampled transition, $a'\sim\pi_o(\cdot\mid s')$, and $\delta$ indicates episode termination. The functions $Q_{\phi_{R,j}}$ and $Q_{\phi_{C,j}}$, $j\in\{1,2\}$, denote the reward and cost critics, respectively, and an overbar denotes the corresponding target-network parameters.

\textbf{Policy and temperature updates.} The online policy minimizes
\begin{equation}
\mathcal L_{\pi_o}
=
\frac{1}{1+\lambda}
\mathbb E_{\substack{s\sim\mathcal B\\a\sim\pi_o(\cdot\mid s)}}
\left[
\alpha\log\pi_o(a\mid s)
-\min_{j\in\{1,2\}}Q_{\phi_{R,j}}(s,a)
+\lambda\max_{j\in\{1,2\}}Q_{\phi_{C,j}}(s,a)
\right].
\label{eq:sac_policy_loss}
\end{equation}
where $s$ is sampled from replay, $a$ is a newly sampled online action, and $\lambda\geq0$ is the Lagrange multiplier. Critic parameters, $\alpha$, and $\lambda$ are held fixed during this update. We adjust $\alpha$ by optimizing $\log\alpha$ with a target entropy equal to the negative action dimension.

\textbf{Multiplier and target-network updates.} Once learning has started, each collection with completed episodes is followed by one Adam step minimizing
\begin{equation}
\mathcal L_{\lambda}=-\lambda\bigl(\bar C_{\mathrm{ep}}-d\bigr),
\label{eq:sac_multiplier_loss}
\end{equation}
where $\bar C_{\mathrm{ep}}$ is the mean cumulative cost of the episodes completed in that collection. We clip $\lambda$ to $[0,5]$ after this update. Each training iteration updates the reward critics, cost critics, online policy, and entropy temperature, then soft-updates the target parameters with weight $0.005$ on the current parameters and $0.995$ on their previous target values.

\subsection{Auxiliary Cost-to-Go Estimator}
\label{app:aux_cost_estimator}

\textbf{Candidate evaluation.} We train two auxiliary cost estimators to estimate the cost-to-go in Eq.~(\ref{eq:cost_to_go}). Both take the state and action as input. Arbitration uses the non-negative mean of their target predictions:
\begin{equation}
\widehat Q_C^{\pi_o}(s,a)
=\left[\frac{1}{2}\sum_{j=1}^{2}Q_{\bar\psi_j}(s,a)\right]_+,
\label{eq:auxiliary_cost_estimate}
\end{equation}
where $Q_{\bar\psi_j}$ denotes the target network of the $j$-th auxiliary cost critic, and $[\cdot]_+$ clips negative predictions to zero.

\textbf{Estimator update.} The auxiliary cost estimators are trained using the TD target
\begin{equation}
y^{\mathrm{aux}}
=c+\gamma(1-\delta)\widehat Q_C^{\pi_o}(s',a'),
\qquad a'\sim\pi_o(\cdot\mid s'),
\label{eq:auxiliary_td_target}
\end{equation}
where $(s,a,c,s',\delta)$ is a transition sampled from $\mathcal B$, and $\gamma$ is the discount factor used in Eq.~(\ref{eq:cost_to_go}). The target estimates the cost of executing the stored action and following the current online policy thereafter. The auxiliary estimators are optimized using the Huber loss, with the remaining training details provided in Table~\ref{tab:rbga_hyperparameters}.

\textbf{Warm-up.} The auxiliary estimator is trained during the first 2,000 online training iterations, while actions are chosen using the guide horizon without budget arbitration. Once both the online learner and auxiliary estimator have completed 2,000 updates, action selection follows Eq.~(\ref{eq:rbga_selection}).

\begin{figure}[t]
    \centering
    \includegraphics[width=0.92\linewidth]{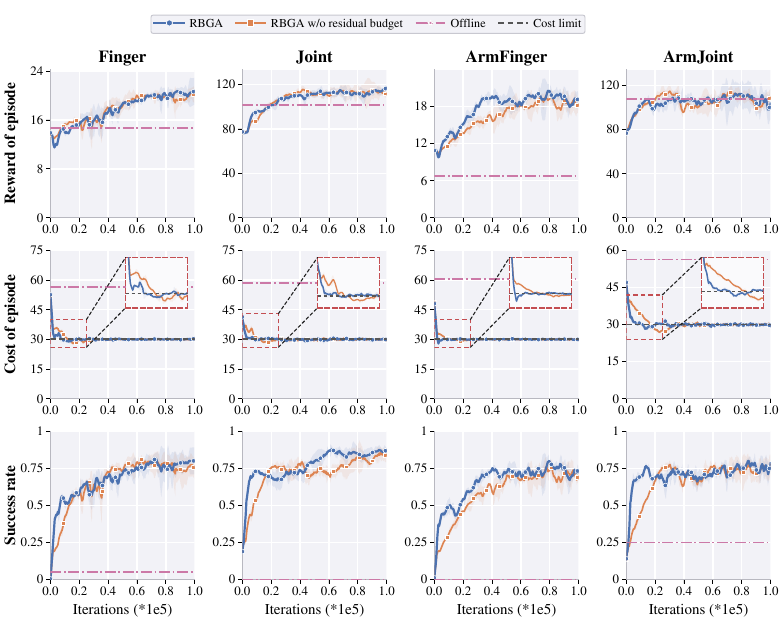}
    \caption{The figure depicts the reward, cost, and success rate curves of RBGA and RBGA without residual-budget arbitration on four tasks. The shaded areas on the curves represent the variance obtained from online testing conducted with three random seeds. The dashed line indicates the episode cost limit $d=30$.}
    \label{fig:ablation_residual_budget}
\end{figure}

\subsection{Guide Horizon Schedules}
\label{app:guide_schedule}

The linear schedule is defined in Eq.~(\ref{eq:linear_guide_schedule}) in Section~\ref{sec:method}. For the step schedule, we predefine decreasing horizon values from $H_0$ to zero and assign them to approximately evenly spaced training stages over the first $N_g$ iterations. The value of $H_i$ is constant within each stage. By default, the horizon values are obtained by evenly spacing values between $H_0$ and zero. Both schedules control guide participation without changing the task horizon $H$.

\section{Additional Experimental Results}
\label{app:additional_results}

\subsection{Effect of Residual-Budget Arbitration}
\label{app:residual_budget_ablation}

To evaluate the contribution of residual-budget arbitration, we disable the residual-budget mechanism while keeping the remaining settings unchanged. As shown in Figure~\ref{fig:ablation_residual_budget}, RBGA satisfies the cost constraint earlier without degrading reward. These results indicate that residual-budget arbitration achieves a more favorable safety--performance trade-off during the early stage of online training. This suggests that incorporating the remaining episode budget helps RBGA select actions with lower predicted cost-to-go while preserving task performance.

\begin{figure}[htbp]
    \centering
    \includegraphics[width=\linewidth]{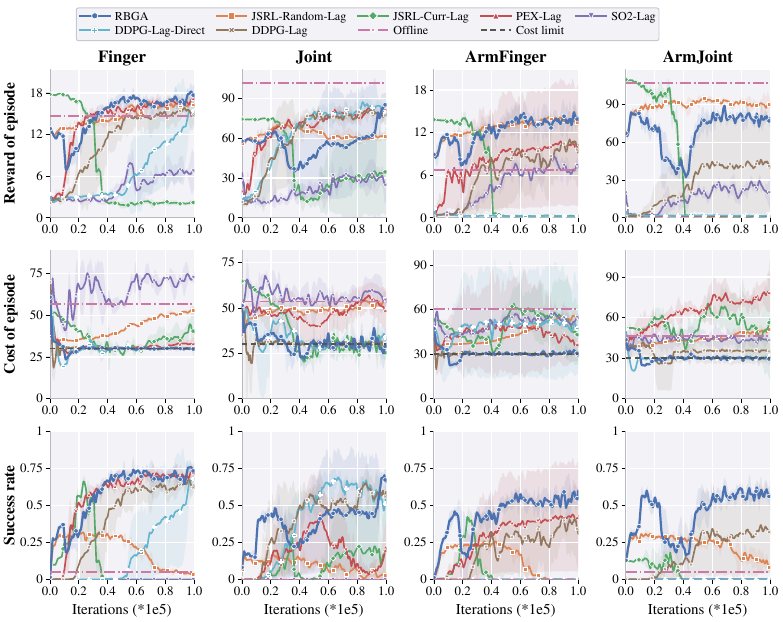}
    \caption{The figure depicts the reward, cost, and success rate curves of RBGA and the baseline algorithms using DDPG-Lag as the online learner. The shaded areas on the curves represent the variance obtained from online testing conducted with three random seeds. The dashed line indicates the episode cost limit $d=30$.}
    \label{fig:ddpg_results}
\end{figure}

\subsection{Results with a DDPG-Lag Online Learner}
\label{app:ddpg_results}

To evaluate the performance of RBGA with a different online RL algorithm, we conduct comparative experiments using DDPG-Lag as the online learner. Specifically, we replace SAC-Lag with DDPG-Lag for RBGA and all corresponding baselines while retaining their original offline-to-online mechanisms. The remaining data collection, guide scheduling, and arbitration settings are kept unchanged.

Figure~\ref{fig:ddpg_results} illustrates the reward, cost, and success rate curves of RBGA and the baseline algorithms on the four tasks. From the illustrated results, it can be observed that RBGA maintains the episode cost close to the predefined cost limit while achieving competitive reward and success rate across the four tasks. Furthermore, RBGA demonstrates stable policy improvement during online training, whereas several baseline algorithms exhibit higher costs or lower task performance. Although the reward of RBGA temporarily decreases during the early stage of training on \emph{ArmJoint}, it subsequently recovers while maintaining a relatively low episode cost and a high success rate. The above results indicate that RBGA can be effectively combined with different constrained online RL algorithms.

\begin{figure}[htbp]
    \centering
    \includegraphics[width=\linewidth]{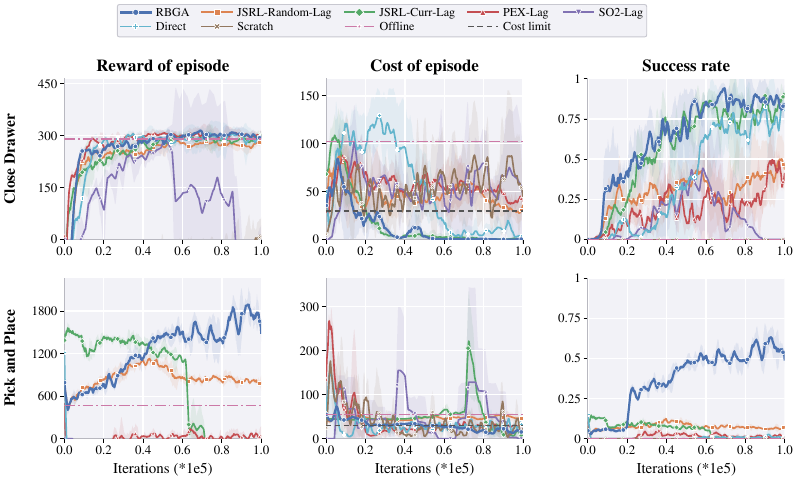}
    \caption{The figure depicts the reward, cost, and success rate curves of RBGA and the baseline algorithms on the \emph{Close Drawer} and \emph{Pick and Place} tasks. The shaded areas on the curves represent the variance obtained from online testing conducted with three random seeds. The dashed line indicates the episode cost limit $d=30$.}
    \label{fig:additional_tasks}
\end{figure}

\subsection{Results on Additional Manipulation Tasks}
\label{app:additional_tasks}

We further evaluate RBGA on two additional manipulation tasks including \emph{Close Drawer} and \emph{Pick and Place}, two Safety-Gymnasium manipulation tasks from Safe Policy Optimization~\citep{ji2023safety}. Both use a mobile base equipped with a Franka arm and a gripper. \emph{Close Drawer} requires the robot to close a drawer while keeping the base outside a restricted region near the cabinet, whereas \emph{Pick and Place} requires the robot to move an object from a lower platform to a higher location while avoiding the restricted region between the platforms. RBGA and all six baselines use SAC-Lag for online learning.

Figure~\ref{fig:additional_tasks} illustrates the reward, cost, and success rate curves of RBGA and the baseline algorithms on the two tasks. From the illustrated results, it can be observed that RBGA consistently improves task performance during online training while maintaining the episode cost within or close to the predefined cost limit. On \emph{Close Drawer}, RBGA achieves high reward and success rate while reducing the episode cost toward the cost limit, whereas several baseline algorithms exhibit higher costs or lower success rates during training. On \emph{Pick and Place}, RBGA continues to improve beyond the offline policy and achieves higher reward and success rate than the compared methods, while several baselines show limited improvement or performance degradation. The above results indicate that RBGA can effectively leverage offline guidance for safe online policy improvement across different manipulation tasks.

\subsection{Real-world Deployment}
\label{app:real_world}

To further evaluate the practical applicability of RBGA, we deploy the policy trained in simulation environment directly on a real-world robotic platform consisting of a RealMan RM65 robotic arm and a Linker Hand L20 dexterous hand. We consider a \textit{Cube Grasping} task, where the same arm--hand platform is modeled in simulation for policy training. For real-world execution, an external Intel RealSense D435 RGB-D camera captures the scene, and FoundationPose~\citep{wen2024foundationpose} estimates the 6D pose of the target object. The estimated object pose is transformed into the robot reference frame and combined with the robot proprioception to construct the same policy observation used during IsaacGym training. The resulting observation is directly fed into the learned policy to generate control commands for the robotic arm and dexterous hand.

\begin{figure}[htbp]
    \centering
    \includegraphics[width=\linewidth]{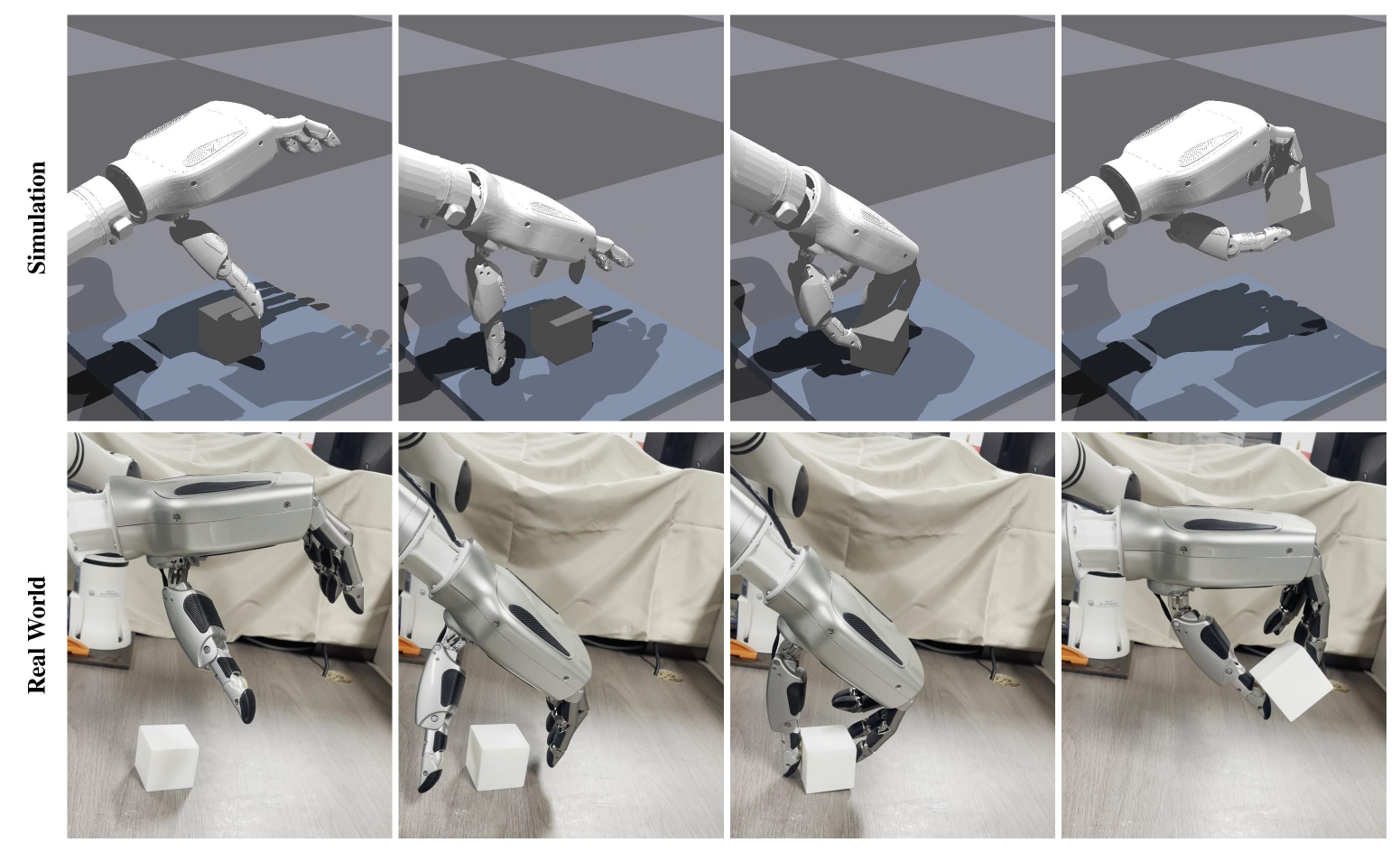}
    \caption{Simulation and real-world deployment of RBGA on the \textit{Cube Grasping} task. Top: simulation. Bottom: the physical robot. The platform consists of a RealMan RM65 robotic arm equipped with a Linker Hand L20 dexterous hand.}
    \label{fig:real_world}
\end{figure}

As shown in Figure~\ref{fig:real_world}, we present the corresponding grasping behaviors in simulation and on the physical robot. The policy learned with RBGA can be transferred to the physical platform and successfully execute the cube grasping behavior while maintaining stable and safety-aware manipulation. This result demonstrates the potential of RBGA for transferring safe dexterous manipulation policies from simulation to real robotic systems.

\section{Proofs and Discussions}
\label{app:proofs}

We provide the proofs of the cost and reward bounds in Section~\ref{sec:theoretical_analysis}. We first derive the Bellman recursion used in Section~\ref{sec:rbga}, and then prove the local and episode cost bounds. Subsequently, we discuss the state coverage and policy advantage conditions and prove the reward performance bound.

\subsection{Proof of the Cost Bellman Recursion}
\label{app:cost_recursion}

For a fixed online policy $\pi_o$, the cost state values at $t=0,\ldots,H-1$ and the terminal value are given as:
\begin{equation}
V_{C}^{\pi_o}(s_t)=\mathbb E_{a_t^o\sim\pi_o(\cdot\mid s_t)}[Q_{C}^{\pi_o}(s_t,a_t^o)], \qquad V_{C}^{\pi_o}(s_H)=0,
\label{eq:app_cost_value}
\end{equation}
where $V_{C}^{\pi_o}(s_t)$ denotes the expected discounted cost-to-go from state $s_t$ at time $t$ under $\pi_o$. The cost-to-go in Eq.~(\ref{eq:cost_to_go}) satisfies the following Bellman recursion:
\begin{equation}
Q_{C}^{\pi_o}(s_t,a_t) =c(s_t,a_t)+\gamma\mathbb E_{s_{t+1}\sim P(\cdot\mid s_t,a_t)}[V_{C}^{\pi_o}(s_{t+1})].
\label{eq:cost_bellman_recursion}
\end{equation}

\begin{proof}
By separating the immediate cost from the remaining discounted cost in Eq.~(\ref{eq:cost_to_go}), we obtain:
\begin{equation}
\begin{aligned}
Q_{C}^{\pi_o}(s_t,a_t) &=c(s_t,a_t)+\gamma\mathbb E_{\pi_o}\left[ \sum_{j=t+1}^{H-1}\gamma^{j-(t+1)}c_j \,\middle|\,s_t,a_t\right]\\
&=c(s_t,a_t)+\gamma\mathbb E_{s_{t+1}\sim P(\cdot\mid s_t,a_t)} \left[V_{C}^{\pi_o}(s_{t+1})\right],
\end{aligned}
\end{equation}
The first equality uses $c_t=c(s_t,a_t)$, and the second equality follows by conditioning on the next state $s_{t+1}$ and using Eq.~(\ref{eq:app_cost_value}). At $t=H-1$, the remaining sum is empty and $V_{C}^{\pi_o}(s_H)=0$, so the same equality holds. Therefore, the cost-to-go satisfies Eq.~(\ref{eq:cost_bellman_recursion}).
\end{proof}

\subsection{Proof and Discussion of Proposition~\ref{prop:local_cost_bounds}}
\label{app:proof_local_cost_bounds}

\noindent \textbf{Assumption~\ref{ass:cost_error}.} 
For a fixed online policy $\pi_o$ and cost estimator $\widehat Q_C^{\pi_o}$, there exists a finite constant $\varepsilon_C\geq0$ such that, at each time step $t<\min\{h,H_i\}$, the cost estimates of both candidate actions in Eq.~(\ref{eq:rbga_selection}) satisfy:
\begin{equation}
\left|\widehat Q_C^{\pi_o}(s_t,a)-Q_{C}^{\pi_o}(s_t,a)\right|\leq\varepsilon_C,\qquad\forall a\in\{a_t^g,a_t^o\},
\label{eq:cost_error_bound_appendix}
\end{equation}
where $a$ ranges over the guide and online candidates, and $\varepsilon_C$ is a uniform bound on their estimation errors.

\noindent\textbf{Proposition~\ref{prop:local_cost_bounds}.}
Under Assumption~\ref{ass:cost_error}, consider a time step $t<\min\{h,H_i\}$. For a selected action satisfying $\gamma^t\widehat Q_C^{\pi_o}(s_t,a_t)\leq b_t$, the accumulated discounted cost and expected continuation cost satisfy:
\begin{equation}
\sum_{j=0}^{t-1}\gamma^j c_j+\gamma^t Q_{C}^{\pi_o}(s_t,a_t) \leq d+\gamma^t\varepsilon_C\leq d+\varepsilon_C.
\end{equation}
For every guide action selected by Eq.~(\ref{eq:rbga_selection}), the difference in cost-to-go satisfies:
\begin{equation}
Q_{C}^{\pi_o}(s_t,a_t^g)-Q_{C}^{\pi_o}(s_t,a_t^o) \leq\zeta+2\varepsilon_C.
\end{equation}

\begin{proof}
By applying the estimation error bound to the selected action and substituting the budget condition, we obtain:
\begin{equation}
\begin{aligned}
\sum_{j=0}^{t-1}\gamma^j c_j+\gamma^t Q_{C}^{\pi_o}(s_t,a_t) &\leq\sum_{j=0}^{t-1}\gamma^j c_j +\gamma^t\widehat Q_C^{\pi_o}(s_t,a_t)+\gamma^t\varepsilon_C\\
&\leq\sum_{j=0}^{t-1}\gamma^j c_j+b_t+\gamma^t\varepsilon_C\\
&=d+\gamma^t\varepsilon_C\leq d+\varepsilon_C,
\end{aligned}
\end{equation}
where the first inequality follows from $Q_{C}^{\pi_o}(s_t,a_t)\leq\widehat Q_C^{\pi_o}(s_t,a_t)+\varepsilon_C$ in Assumption~\ref{ass:cost_error}, the second uses $\gamma^t\widehat Q_C^{\pi_o}(s_t,a_t)\leq b_t$, the equality follows from Eq.~(\ref{eq:remaining_budget}), and the final inequality uses $\gamma^t\leq1$. Therefore, Eq.~(\ref{eq:local_budget_bound}) holds.

To derive the guide comparison bound, we consider the branches of Eq.~(\ref{eq:rbga_selection}) that select a guide action. If the selected guide action passes the budget condition, the second branch of the selection rule gives:
\begin{equation}
\widehat q_t^g-\widehat q_t^o \leq\min\{\gamma^t\zeta,b_t-\widehat q_t^o\}.
\end{equation}
This inequality covers the cases in which both candidates pass the budget condition and in which only the guide candidate passes. If the selected guide action fails the budget condition, the online candidate also fails, and the minimum-cost fallback gives $\widehat q_t^g-\widehat q_t^o\leq0$. Since $\gamma^t\zeta\geq0$, the estimate difference in both cases satisfies:
\begin{equation}
\widehat q_t^g-\widehat q_t^o \leq\min\{\gamma^t\zeta,[b_t-\widehat q_t^o]_+\}.
\end{equation}
where $[b_t-\widehat q_t^o]_+$ is the non-negative budget remaining after accounting for the online candidate's estimated discounted cost.

By Assumption~\ref{ass:cost_error}, the guide candidate satisfies $Q_{C}^{\pi_o}(s_t,a_t^g)\leq\widehat Q_C^{\pi_o}(s_t,a_t^g)+\varepsilon_C$, while the online candidate satisfies $Q_{C}^{\pi_o}(s_t,a_t^o)\geq\widehat Q_C^{\pi_o}(s_t,a_t^o)-\varepsilon_C$. Applying these two bounds to the cost-to-go difference gives:
\begin{equation}
\begin{aligned}
&\gamma^t\left[Q_{C}^{\pi_o}(s_t,a_t^g)-Q_{C}^{\pi_o}(s_t,a_t^o)\right]\\
&\quad\leq\gamma^t\left[\widehat Q_C^{\pi_o}(s_t,a_t^g)+\varepsilon_C-\left(\widehat Q_C^{\pi_o}(s_t,a_t^o)-\varepsilon_C\right)\right]\\
&\quad=\gamma^t\left[\widehat Q_C^{\pi_o}(s_t,a_t^g)-\widehat Q_C^{\pi_o}(s_t,a_t^o)\right]+2\gamma^t\varepsilon_C\\
&\quad=\widehat q_t^g-\widehat q_t^o+2\gamma^t\varepsilon_C\\
&\quad\leq2\gamma^t\varepsilon_C+\min\{\gamma^t\zeta,[b_t-\widehat q_t^o]_+\}\\
&\quad\leq\gamma^t(\zeta+2\varepsilon_C).
\end{aligned}
\label{eq:budget_guide_cost_bound}
\end{equation}
The second equality uses the definitions of $\widehat q_t^g$ and $\widehat q_t^o$. The second inequality follows from the preceding bound on their difference, and the final inequality uses $\min\{\gamma^t\zeta,[b_t-\widehat q_t^o]_+\}\leq\gamma^t\zeta$. Dividing both sides by $\gamma^t>0$ yields Eq.~(\ref{eq:guide_selection_cost_bound}).

\end{proof}

\subsection{Proof and Discussion of Proposition~\ref{thm:global_cost_bound}}
\label{app:proof_global_cost_bound}

\noindent\textbf{Proposition~\ref{thm:global_cost_bound}.}
Under Assumption~\ref{ass:cost_error}, fix $\pi_o$, $\widehat Q_C^{\pi_o}$, and $H_i$ throughout an episode, and sample $h\sim\operatorname{Unif}\{0,\ldots,H_i\}$ independently of the initial state. The composite policy $\pi$ satisfies:
\begin{equation}
[J_C(\pi)-d]_+ \leq[J_C(\pi_o)-d]_+ +(\zeta+2\varepsilon_C) \sum_{t=0}^{H_i-1}\gamma^t\frac{H_i-t}{H_i+1}.
\end{equation}

\begin{proof}
The composite policy $\pi$ may depend on the interaction history through $b_t$ and on the sampled horizon $h$, while the online candidate is sampled from $\pi_o(\cdot\mid s_t)$.

To compare the expected discounted costs of $\pi$ and $\pi_o$, we use the finite-horizon cost form of the performance difference lemma \citep{kakade2002approximately}. With both policies starting from $s_0\sim\rho_0$, we obtain:
\begin{equation}
J_C(\pi)-J_C(\pi_o)=\sum_{t=0}^{H-1}\gamma^t\mathbb E_{\pi}\left[Q_{C}^{\pi_o}(s_t,a_t)-V_{C}^{\pi_o}(s_t)\right],
\label{eq:finite_horizon_cost_pdl}
\end{equation}
where the expectation is over trajectories generated by $\pi$, including the sampled guide horizon and candidate actions. Taking expectations in Eq.~(\ref{eq:cost_bellman_recursion}) and summing the discounted differences cancels the intermediate value terms. Using $V_C^{\pi_o}(s_H)=0$ and $\mathbb E_{s_0\sim\rho_0}[V_C^{\pi_o}(s_0)]=J_C(\pi_o)$ then gives Eq.~(\ref{eq:finite_horizon_cost_pdl}).

For RBGA, the online candidate is sampled as $a_t^o\sim\pi_o(\cdot\mid s_t)$. Conditional on the history before candidate sampling, its expected cost-to-go equals $V_{C}^{\pi_o}(s_t)$ by Eq.~(\ref{eq:app_cost_value}). Substituting this relation into Eq.~(\ref{eq:finite_horizon_cost_pdl}) gives:
\begin{equation}
J_C(\pi)-J_C(\pi_o)=\sum_{t=0}^{H-1}\gamma^t\mathbb E_{\pi}\left[Q_{C}^{\pi_o}(s_t,a_t)-Q_{C}^{\pi_o}(s_t,a_t^o)\right].
\label{eq:cost_difference_identity}
\end{equation}

When the online candidate is selected, the action-value difference is zero. When the guide candidate is selected, Eq.~(\ref{eq:guide_selection_cost_bound}) bounds this difference by $\zeta+2\varepsilon_C$. By substituting these bounds into Eq.~(\ref{eq:cost_difference_identity}), we obtain:
\begin{equation}
J_C(\pi)-J_C(\pi_o) \leq(\zeta+2\varepsilon_C)\sum_{t=0}^{H-1}\gamma^t \Pr_{\pi}(I_t^g=1),
\label{eq:global_expected_cost_bound}
\end{equation}
where $I_t^g\in\{0,1\}$ equals one when the guide candidate is selected under Eq.~(\ref{eq:rbga_selection}) before episode termination, including selection by the fallback, and zero otherwise. The probability $\Pr_{\pi}(I_t^g=1)$ includes the randomness of the trajectory, candidate actions, and sampled horizon. Since guide selection requires $t<h$ and $h$ is uniform on $\{0,\ldots,H_i\}$, the discounted sum of these probabilities satisfies:
\begin{equation}
\sum_{t=0}^{H-1}\gamma^t \Pr_{\pi}(I_t^g=1) \leq\sum_{t=0}^{H-1}\gamma^t\Pr(h>t) =\sum_{t=0}^{H_i-1}\gamma^t\frac{H_i-t}{H_i+1}.
\end{equation}

Combining this inequality with Eq.~(\ref{eq:global_expected_cost_bound}), taking positive parts, and using $[u+v]_+\leq[u]_++v$ for $v\geq0$ completes the proof:
\begin{equation}
\begin{aligned}
[J_C(\pi)-d]_+
&\leq\left[J_C(\pi_o)-d+(\zeta+2\varepsilon_C)
\sum_{t=0}^{H_i-1}\gamma^t\frac{H_i-t}{H_i+1}\right]_+\\
&\leq[J_C(\pi_o)-d]_++(\zeta+2\varepsilon_C)
\sum_{t=0}^{H_i-1}\gamma^t\frac{H_i-t}{H_i+1}.
\end{aligned}
\end{equation}
\end{proof}

Proposition~\ref{thm:global_cost_bound} establishes the episode cost bound with the online policy, cost estimator, and maximum guide horizon fixed throughout the episode. In online learning, an episode may instead span multiple policy updates. These updates can change the expected remaining cost, so their effect must be included when bounding the cost of the resulting interaction. Proposition~\ref{prop:changing_policy_cost_bound} extends the analysis to this setting by adding a term for increases in the online policy's cost value. Under the fixed-policy conditions of Proposition~\ref{thm:global_cost_bound}, this term vanishes and the original bound is recovered.

\begin{proposition}[Cost bound with policy updates]
\label{prop:changing_policy_cost_bound}
Let $i_t$ be the current training iteration before action selection at episode step $t$, and let $\pi_o^{(t)}$ denote the online policy available at that step. The superscript $(t)$ distinguishes policy versions only in this extension. Fix the initial learner state, including $i_0$ and $\pi_o^{(0)}$, before sampling $s_0\sim\rho_0$ and $h\sim\operatorname{Unif}\{0,\ldots,H_{i_0}\}$ independently. Keep $h$ fixed throughout the episode. At each step, the current online policy and cost estimator are determined before candidate sampling. Assume Eq.~(\ref{eq:cost_error_bound}) holds with the same $\varepsilon_C$ for both candidates at every step satisfying $t<\min\{h,H_{i_t}\}$. If Eq.~(\ref{eq:rbga_selection}) is used throughout the episode with the current policy, estimator, and maximum guide horizon, then:
\begin{equation}
\begin{aligned}
\mathbb E\left[\sum_{t=0}^{H-1}\gamma^t c_t\right]
&\leq J_C(\pi_o^{(0)})\\
&\quad+\mathbb E\left[\sum_{t=0}^{H-1}I_t^g\left(2\gamma^t\varepsilon_C+\min\{\gamma^t\zeta,[b_t-\widehat q_t^o]_+\}\right)\right]\\
&\quad+\sum_{t=1}^{H-1}\gamma^t\mathbb E\left[\left[V_C^{\pi_o^{(t)}}(s_t)-V_C^{\pi_o^{(t-1)}}(s_t)\right]_+\right],
\end{aligned}
\label{eq:changing_policy_cost_bound}
\end{equation}
where $\widehat q_t^o=\gamma^t\widehat Q_C^{\pi_o^{(t)}}(s_t,a_t^o)$ uses the current cost estimator, and $I_t^g$ indicates guide selection. The expectations include randomness in interaction and policy updates, conditional on the initial learner state. Both value functions in the last term are evaluated at the same state $s_t$ and episode step $t$. For episodes that terminate before $H$, remaining costs and terminal-state values are set to zero.
\end{proposition}


\begin{proof}
Applying the Bellman recursion in Eq.~(\ref{eq:cost_bellman_recursion}) to the current policy, held fixed for each value evaluation, and taking expectations gives:
\begin{equation}
\begin{aligned}
&\mathbb E\left[Q_C^{\pi_o^{(t)}}(s_t,a_t)-V_C^{\pi_o^{(t)}}(s_t)\right]\\
&\quad=\mathbb E\left[c_t+\gamma V_C^{\pi_o^{(t)}}(s_{t+1})-V_C^{\pi_o^{(t)}}(s_t)\right].
\end{aligned}
\end{equation}
Multiplying by $\gamma^t$ and summing over $t$ leaves the value differences between successive policies at the intermediate states. Moreover, conditional on the history before candidate sampling, $a_t^o\sim\pi_o^{(t)}(\cdot\mid s_t)$ implies that its expected action value equals $V_C^{\pi_o^{(t)}}(s_t)$. Using the common initial distribution and zero terminal values therefore yields:
\begin{equation}
\begin{aligned}
&\mathbb E\left[\sum_{t=0}^{H-1}\gamma^t c_t\right]-J_C(\pi_o^{(0)})\\
&\quad=\sum_{t=0}^{H-1}\gamma^t\mathbb E\left[Q_C^{\pi_o^{(t)}}(s_t,a_t)-Q_C^{\pi_o^{(t)}}(s_t,a_t^o)\right]\\
&\qquad+\sum_{t=1}^{H-1}\gamma^t\mathbb E\left[V_C^{\pi_o^{(t)}}(s_t)-V_C^{\pi_o^{(t-1)}}(s_t)\right].
\end{aligned}
\label{eq:changing_policy_identity}
\end{equation}
The action-value difference is zero when the online candidate is selected. For a selected guide candidate, Eq.~(\ref{eq:budget_guide_cost_bound}) applies to the current policy and estimator under the stated error condition. Applying this bound to the first sum and replacing each policy value difference in the second sum by its positive part proves Eq.~(\ref{eq:changing_policy_cost_bound}).
\end{proof}

Even when policies and guide limits change within an episode, guide selection requires $t<h$. The guide contribution can therefore be bounded using the horizon sampled at episode start:
\begin{equation}
\begin{aligned}
&\mathbb E\left[\sum_{t=0}^{H-1}I_t^g\left(2\gamma^t\varepsilon_C+\min\{\gamma^t\zeta,[b_t-\widehat q_t^o]_+\}\right)\right]\\
&\quad\leq(\zeta+2\varepsilon_C)\sum_{t=0}^{H-1}\gamma^t\Pr(h>t)\\
&\quad=(\zeta+2\varepsilon_C)\sum_{t=0}^{H_{i_0}-1}\gamma^t\frac{H_{i_0}-t}{H_{i_0}+1}.
\end{aligned}
\label{eq:changing_policy_guide_bound}
\end{equation}
Taking positive parts in Eq.~(\ref{eq:changing_policy_cost_bound}) then gives:
\begin{equation}
\begin{aligned}
&\left[\mathbb E\left[\sum_{t=0}^{H-1}\gamma^t c_t\right]-d\right]_+\\
&\quad\leq[J_C(\pi_o^{(0)})-d]_++(\zeta+2\varepsilon_C)\sum_{t=0}^{H_{i_0}-1}\gamma^t\frac{H_{i_0}-t}{H_{i_0}+1}\\
&\qquad+\sum_{t=1}^{H-1}\gamma^t\mathbb E\left[\left[V_C^{\pi_o^{(t)}}(s_t)-V_C^{\pi_o^{(t-1)}}(s_t)\right]_+\right].
\end{aligned}
\label{eq:changing_policy_violation_bound}
\end{equation}

The final term accounts for increases in cost value caused by policy updates. It vanishes when the policy is unchanged or each update satisfies $V_C^{\pi_o^{(t)}}(s_t)\leq V_C^{\pi_o^{(t-1)}}(s_t)$ almost surely at the visited state and episode step. In particular, under the fixed-policy conditions of Proposition~\ref{thm:global_cost_bound}, Eq.~(\ref{eq:changing_policy_violation_bound}) reduces to its bound. For general updates, Eq.~(\ref{eq:changing_policy_cost_bound}) guarantees that the expected episode cost does not exceed $d$ if its right-hand side is at most $d$. Assumption~\ref{ass:cost_error} alone does not control the changes in policy value.

\subsection{Discussion of Assumption~\ref{ass:rollin_learning}}
\label{app:policy_search_rollins}

Assumption~\ref{ass:rollin_learning} gives conditions on the state distributions generated by RBGA roll-ins and the reward advantage of the learned policy. The distribution $\mu_t$ is induced by RBGA roll-ins, whose action selections may depend on the remaining budget, sampled horizon, and earlier observations. The value $Q_{R}^{\pi_o}$ evaluates the action at time $t$ followed by the learned online policy $\pi_o$, as in guide-based policy search \citep[Appendix A.5]{uchendu2023jumpstart}.

The state coverage condition in Eq.~(\ref{eq:rbga_rollin_coverage}) compares probabilities of measurable sets. In a finite state space, it gives $\rho_t^{\pi^\star}(s)\leq D_\infty\mu_t(s)$ for every state. Under this condition, the expectation of any non-negative measurable function satisfies:
\begin{equation}
\int f(s)\,\rho_t^{\pi^\star}(\mathrm ds) \leq D_\infty\int f(s)\,\mu_t(\mathrm ds),
\end{equation}
where $f$ is a non-negative measurable function of the state. The constant $D_\infty$ bounds how much more frequently the optimal feasible policy visits a set of states than the roll-in policy. Since the remaining budget affects action selection, the coverage condition is imposed on the distributions generated by the complete RBGA selection rule.

The advantage condition in Eq.~(\ref{eq:rollin_advantage_error}) takes the positive part after averaging over the comparator's actions and before averaging over states. This gives the non-negative function to which the state coverage inequality applies. For the optimal feasible policy itself, the expected advantage satisfies:
\begin{equation}
\mathbb E_{a\sim\pi^\star(\cdot\mid s)}A_{R}^{\pi^\star}(s,a)=0.
\end{equation}

\subsection{Proof and Discussion of Proposition~\ref{thm:screened_exploration}}
\label{app:proof_reward_bound}

\noindent\textbf{Assumption~\ref{ass:rollin_learning}.}
For the learned online policy $\pi_o$, there exist a finite coverage coefficient $D_\infty\geq1$ and finite reward advantage bounds $\varepsilon_{R,t}\geq0$ such that, at every time step $t$, the state distributions satisfy:
\begin{equation}
\rho_t^{\pi^\star}\leq D_\infty\mu_t,
\label{eq:rbga_rollin_coverage_appendix}
\end{equation}
where $\mu_t$ is the marginal state distribution at time $t$ induced by RBGA roll-ins, $\rho_t^{\pi^\star}$ is the corresponding distribution under $\pi^\star$, and $D_\infty$ bounds their relative probabilities. Under the roll-in distribution, the expected positive advantage of the optimal feasible policy satisfies:
\begin{equation}
\mathbb E_{s\sim\mu_t}
\left[\left[\mathbb E_{a\sim\pi^\star(\cdot\mid s)}
A_{R}^{\pi_o}(s,a)\right]_+\right]
\leq\varepsilon_{R,t},
\label{eq:rollin_advantage_error_appendix}
\end{equation}
where $\varepsilon_{R,t}$ bounds the remaining positive reward advantage of $\pi^\star$ under the roll-in distribution.

\noindent\textbf{Proposition~\ref{thm:screened_exploration}.}
Under Assumption~\ref{ass:rollin_learning}, the reward gap between the learned online policy $\pi_o$ and the optimal feasible policy $\pi^\star$ satisfies:
\begin{equation}
J_R(\pi^\star)-J_R(\pi_o) \leq D_\infty\sum_{t=0}^{H-1}\gamma^t\varepsilon_{R,t}.
\end{equation}

\begin{proof}
To compare the reward performance of $\pi^\star$ and $\pi_o$, we use the finite-horizon reward form of the performance difference lemma \citep[Lemma~6.1]{kakade2002approximately}. With both policies starting from $\rho_0$, we obtain:
\begin{equation}
J_R(\pi^\star)-J_R(\pi_o) =\sum_{t=0}^{H-1}\gamma^t \mathbb E_{s\sim\rho_t^{\pi^\star}} \left[\mathbb E_{a\sim\pi^\star(\cdot\mid s)}A_{R}^{\pi_o}(s,a)\right].
\end{equation}

By replacing each state's expected advantage with its positive part and applying the two conditions in Assumption~\ref{ass:rollin_learning}, we obtain:
\begin{equation}
\begin{aligned}
J_R(\pi^\star)-J_R(\pi_o) &\leq\sum_{t=0}^{H-1}\gamma^t \mathbb E_{s\sim\rho_t^{\pi^\star}} \left[\left[\mathbb E_{a\sim\pi^\star(\cdot\mid s)}A_{R}^{\pi_o}(s,a)\right]_+\right]\\
&\leq D_\infty\sum_{t=0}^{H-1}\gamma^t \mathbb E_{s\sim\mu_t} \left[\left[\mathbb E_{a\sim\pi^\star(\cdot\mid s)}A_{R}^{\pi_o}(s,a)\right]_+\right]\\
&\leq D_\infty\sum_{t=0}^{H-1}\gamma^t\varepsilon_{R,t},
\end{aligned}
\end{equation}
where the second inequality applies Eq.~(\ref{eq:rbga_rollin_coverage_appendix}) to a non-negative function, and the last inequality follows from Eq.~(\ref{eq:rollin_advantage_error_appendix}). Therefore, the reward gap satisfies Eq.~(\ref{eq:reward_performance_bound}).
\end{proof}

For the same learned policy satisfying $J_C(\pi_o)\leq d$, the optimality of $\pi^\star$ gives:
\begin{equation}
0\leq J_R(\pi^\star)-J_R(\pi_o) \leq D_\infty\sum_{t=0}^{H-1}\gamma^t\varepsilon_{R,t}.
\end{equation}
Thus, for a sequence of feasible learned policies with a common finite coverage constant and $\varepsilon_{R,t}$ tending to zero at every time step, the reward gap tends to zero.

\end{document}

%% file: math_commands.tex
\usepackage{amsmath,amsfonts,bm}

\def\eqref#1{equation~\ref{#1}}

\def\1{\bm{1}}

\DeclareMathAlphabet{\mathsfit}{\encodingdefault}{\sfdefault}{m}{sl}
\SetMathAlphabet{\mathsfit}{bold}{\encodingdefault}{\sfdefault}{bx}{n}

